# Bilinear discriminant feature line analysis for image feature extraction

Lijun Yan, Jun-Bao Li, Xiaorui Zhu, Jeng-Shyang Pan and Linlin Tang✉

A novel bilinear discriminant feature line analysis (BDFLA) is proposed for image feature extraction. The nearest feature line (NFL) is a powerful classifier. Some NFL-based subspace algorithms were introduced recently. In most of the classical NFL-based subspace learning approaches, the input samples are vectors. For image classification tasks, the image samples should be transformed to vectors first. This process induces a high computational complexity and may also lead to loss of the geometric feature of samples. The proposed BDFLA is a matrix-based algorithm. It aims to minimise the within-class scatter and maximise the between-class scatter based on a two-dimensional (2D) NFL. Experimental results on two-image databases confirm the effectiveness.

*Introduction:* The nearest feature line (NFL), proposed by Li and Lu in 1999 [1], is a powerful classifier for image classification. Its kernel is a feature line (FL) metric. It measures the distance between a query sample and some class using the distance between the query sample and the FL of the corresponding class, rather than that between the query sample and the prototype sample in the corresponding class. Some NFL-based subspace learning algorithms were designed for feature extraction, including the NFL space (NFLS) [2], the uncorrelated discriminant NFL analysis (UDNFLA) [3] and so on. However, to use most of current NFL-based feature extraction algorithms, the image samples should be transformed into vectors first. This will increase the computational complexity and may lead to loss of the geometric feature of the image samples. In this Letter, a novel image feature extraction algorithm called bilinear discriminant feature line analysis (BDFLA) is proposed. The proposed BDFLA can extract the feature from the image matrix directly.

*Uncorrelated discriminant NFL analysis:* UDNFLA is a subspace learning method based on NFL. Given a prototype sample set, $\Pi = \{x_1, x_2, \ldots, x_N\} \in R^D$, denote $x^i_{m,n}$ as the project point of sample $x_i$ to the FL $l_{m,n}$ spanned by $x_m$ and $x_n$. The optimisation function of UDNFLA is

$$W = \arg\min \, [\mathrm{tr}(W^T(A-B)W)] \quad \text{subject to } W^T S_t W = I \tag{1}$$

Then the optimisation problem can be transformed to the following eigenvalue problem:

$$(A-B)W = \lambda S_t W \tag{2}$$

where

$$S_t = \frac{1}{N}\sum_{i=1}^{N}(x_i - EX)(x_i - EX)^T \tag{3}$$

$$EX = \frac{1}{N}\sum_{i=1}^{N} x_i \tag{4}$$

$$A = \sum_{i=1}^{N} \frac{1}{NN_i} \sum_{x_m \in P(x_i)} (x_i - x^i_{m,n})(x_i - x^i_{m,n})^T \tag{5}$$

Here, $N_i$ denotes the number of FLs in the same class with $x_i$ and $x_m \in P(x_i)$ means $x_m$ and $x_i$ are in the same class

$$B = \sum_{i=1}^{N} \frac{1}{NM_i} \sum_{x_m \notin P(x_i)} (x_i - x^i_{m,n})(x_i - x^i_{m,n})^T \tag{6}$$

where $M_i$ denotes the number of FLs in the different class with $x_i$ and $x_m \notin P(x_i)$ means $x_m$ and $x_i$ belong to two different classes, respectively.

*Proposed algorithm:* In the classical NFL classifier, the input samples should be vectors. In this Section, a two-dimensional (2D) NFL is presented using the similar idea of NFL. In 2D NFL, all the matrices with the same size are viewed as the points of the linear space. Given two matrices $A = [a_{ij}]$ and $B = [b_{ij}]$, let

$$\|A\| = \sqrt{\sum_{ij}\|a_{ij}\|^2} \quad \text{and} \quad \langle A, B\rangle = \sum_{ij} a_{ij} b_{ij}$$

Given a prototype sample set, $\Pi = \{X_1, X_2, \ldots, X_N\} \in R^{D_1 \times D_2}$, the FL $l_{i,j}$ is as follows:

$$Y = X_i + \mu(X_j - X_i) \tag{7}$$

Using the same method in NFL, the nearest matrix between query sample $Q$ and the FL $l_{i,j}$ is $Q_p = X_1 + \mu_0(X_2 - X_1)$ where

$$\mu_0 = \frac{\langle Q - X_1, X_2 - X_1 \rangle}{\langle X_2 - X_1, X_2 - X_1 \rangle} \tag{8}$$

Let

$$\mathrm{dis}_{2D\,NFL}(Q, l_{i,j}) = \|Q - Q_p\|$$

If there exists an FL $l_{i_0,j_0}$ such that

$$\mathrm{dis}_{2D\,NFL}(Q, i_0, j_0) = \min_{l_{ij}}\{\mathrm{dis}_{2D\,NFL}(Q, l_{i,j})\}$$

then the query image sample $Q$ will be assigned to the class $l(x_{i_0})$.

Given a prototype image sample set $X = X_1, X_2, \ldots, X_N \subset R^{D_1 \times D_2}$, the between-class scatter based on 2D NFL $S_{bFL}$ and the within-class scatter based on 2D NFL $S_{wFL}$ are introduced as follows:

$$S_{bFL} = \frac{1}{N}\sum_{i=1}^{N}\frac{1}{M_i}\sum_{X_m \notin P(X_i)} \|L^T X_i R - L^T X^i_{m,n} R\|^2 \tag{9}$$

and

$$S_{wFL} = \frac{1}{N}\sum_{i=1}^{N}\frac{1}{N_i}\sum_{X_m \in P(X_i)} \|L^T X_i R - L^T X^i_{m,n} R\|^2 \tag{10}$$

where $X^i_{m,n}$ is a matrix, which is also a project point of $X_i$ to the FL generalised by $X_m$ and $X_n$ in the matrix linear space, $N_i$ is the number of FLs in the same class with $X_i$, $l(X_m)$ denotes the class label of $X_m$ and $N_{l(X_m)}$ is the number of FLs in the $l(X_m)$ class. Note that from the definition of 2D NFL, the class label of $X_m$ equals the class label of $X_n$. $S_{wNF}$ computes the square sum of the distances between each prototype sample and the 2D FLs in the same class with the corresponding prototype sample. $S_{bNF}$ calculates the square sum of the distances between each prototype sample and the 2D FLs in the different class. Therefore, $S_{wNF}$ can evaluate the within-class scatter of the prototype image samples and $S_{bNF}$ can measure the between-class scatter of the prototype samples.

The proposed BDFLA aims to minimise the within-class scatter based on 2D NFL and maximise the between-class scatter based on 2D NFL. Therefore, to obtain two optimal maps, $L \in R^{D_1 \times d_1}$ and $R \in R^{D_2 \times d_2}$, the criterion of the proposed algorithm is defined as follows:

$$\max J(L, R) = S_{bFL} - S_{wFL} \tag{11}$$

then

$$\begin{aligned}
S_{wFL} &= \frac{1}{N}\sum_{i=1}^{N}\frac{1}{N_i}\sum_{X_m \in P(X_i)}\|R^T X_i L - R^T X^i_{m,n} L\|^2 \\
&= \sum_{i=1}^{N}\frac{1}{NN_i}\sum_{X_m \in P(X_i)} \mathrm{tr}\big[R^T(X_i - X^i_{m,n})LL^T(X_i - X^i_{m,n})^T R\big] \\
&= \mathrm{tr}\sum_{i=1}^{N}\frac{1}{NN_i}\sum_{X_m \in P(X_i)} \big[R^T(X_i - X^i_{m,n})LL^T(X_i - X^i_{m,n})^T R\big] \\
&= \mathrm{tr}\, R^T\bigg\{\sum_{i=1}^{N}\frac{1}{NN_i}\sum_{X_m \in P(X_i)}\big[(X_i - X^i_{m,n})LL^T(X_i - X^i_{m,n})^T\big]\bigg\}R \\
&= \mathrm{tr}\, R^T S^L_w R
\end{aligned} \tag{12}$$

where tr denotes the trace of a matrix, and

$$S^L_w = \sum_{i=1}^{N}\frac{1}{NN_i}\sum_{X_m \in P(X_i)}\big[(X_i - X^i_{m,n})LL^T(X_i - X^i_{m,n})^T\big] \tag{13}$$

At the same time

$$S_{wFL} = \frac{1}{N} \sum_{i=1}^{N} \frac{1}{N_i} \sum_{X_m \in P(X_i)} \|R^T X_i L - R^T X_{m,n}^i L\|^2$$

$$= \sum_{i=1}^{N} \frac{1}{NN_i} \sum_{X_m \in P(X_i)} \text{tr}[L^T (X_i - X_{m,n}^i)^T RR^T (X_i - X_{m,n}^i) L]$$

$$= \text{tr } L^T S_w^R L \quad (14)$$

where

$$S_w^R = \sum_{i=1}^{N} \frac{1}{NN_i} \sum_{X_m \in P(X_i)} [(X_i - X_{m,n}^i)^T RR^T (X_i - X_{m,n}^i) L] \quad (15)$$

Similar to the above matrix computation

$$S_{bFL} = \text{tr}(L^T S_b^R L) = \text{tr}(R^T S_b^L R) \quad (16)$$

where

$$S_b^R = \frac{1}{N} \sum_{i=1}^{N} \frac{1}{M_i} \sum_{X_m \notin P(X_i)} L^T (X_i - X_{m,n}^i) RR^T (X_i - X_{m,n}^i)^T L \quad (17)$$

$$S_b^L = \frac{1}{N} \sum_{i=1}^{N} \frac{1}{M_i} \sum_{X_m \notin P(X_i)} R^T (X_i - X_{m,n}^i) LL^T (X_i - X_{m,n}^i)^T R \quad (18)$$

Finally

$$J(L, R) = \text{tr } L^T (S_b^R - S_w^R) L = \text{tr } R^T (S_b^L - S_w^L) R \quad (19)$$

An iterative procedure is presented to solve the problem in (19). For a given $R^{t-1} \in R^{D_2 \times d_2}$, $J$ can be rewritten as $J_R^t = L^T (S_b^{R^{t-1}} - S_w^{R^{t-1}}) L$. An approximate solution of $L$ can be calculated using eigenvalue decomposition

$$J_R^t l_i^t = \lambda_i l_i^t \quad (20)$$

That is $L^t = [l_1^t, l_2^t, \ldots, l_{d_1}^t]$, where $l_1^t, l_2^t, \ldots, l_{d_1}^t$ are eigenvectors corresponding to $d_1$ biggest eigenvalues of $J_L^t$. Similarly, for a given $L^t \in R^{D_1 \times d_1}$, $J$ can be rewritten as $J_L^t = R^T (S_b^{L^t} - S_w^{L^t}) R$. Using eigenvalue decomposition again, an approximate solution can be obtained

$$J_L^t r_i^t = \lambda_i r_i^t \quad (21)$$

Denote $R^t = [r_1^t, r_2^t, \ldots, r_{d_2}^t]$, where $r_1^t, r_2^t, \ldots, r_{d_2}^t$ are eigenvectors corresponding to $d_2$ biggest eigenvalues of $J_R^t$. The above procedure is repeated to find the final solution. The procedure of the algorithm is as follows.

**Algorithm 1** Proposed BDFLA

**Require:** The prototype image samples $X = \{X_1, X_2, \ldots, X_N\} \subset R^{D_1 \times D_2}$, $d_1$, $d_2$, the iteration number $T_{\text{Max}}$ and the threshold $\varepsilon$.
**Ensure:** $L \in R^{D_1 \times d_1}$ and $R \in R^{D_2 \times d_2}$
  $t \leftarrow 0$
  $R^t \leftarrow I_{D_2}$
  $L^t \leftarrow I_{D_1}$
  while $t < T_{\text{Max}}$ do
    $t \leftarrow t + 1$
    Compute $S_b^{R^{t-1}}$ with (17)
    Compute $S_w^{R^{t-1}}$ with (15)
    Compute the projection matrix $L^t$ by solving (20)
    Compute the projection matrix $R^t$ by solving (21)
    if $\|L^t - L^{t-1}\|^2 + \|R^t - R^{t-1}\|^2 < \varepsilon$ then
      Break
    end if
  end while

Then, for an image sample $I$, $F = L^T I R \in R^{d_1 \times d_2}$ is the feature extracted by BDFLA and is used for classification.

*Experimental results:* In this Section, the COIL20 database [4] and the FKP database [5] are used to evaluate the proposed algorithms. In the following experiments, NFL is used for classification. The system runs 20 times. The average maximum recognition rate (AMRR) with a corresponding feature dimension is given. To evaluate the performance of the proposed algorithms, BDFLA was compared with the principal component analysis (PCA) [6], the linear discriminant analysis (LDA) [7], the 2D-PCA [8], 2D-LDA [7], NFLS, UDNFLA and NFL embedding (NFLE) [9] in the experiments.

To reduce the computation complexity, all the image samples in the COIL20 database were cropped to $48 \times 48$. About 10 image samples per class were selected randomly for training and the rest were for test. For the FKP database, instead of treating each person's fingers as one subject, each finger was treated as one subject in this experiment. Some duplicate samples were removed from the database. Each sample was cropped to $40 \times 60$. Five image samples per class were chosen randomly for training and the other samples were used for test. For vector-based algorithms, PCA was first performed on the FKP database and 97% energy was preserved.

Table 1 shows the experimental results on the COIL20 database and the FKP database. From the Table, BDFLA has higher AMRRs than the other algorithms.

**Table 1:** Experimental results on two databases

| Algorithm | COIL20 database | | FKP database | |
|---|---|---|---|---|
| | AMRR (%) | Dimension | AMRR (%) | Dimension |
| PCA | 85.91 | 100 | 91.59 | 160 |
| LDA | 88.23 | 19 | 93.73 | 190 |
| NFLS | 87.96 | 120 | 90.84 | 160 |
| UDNFLA | 89.32 | 130 | 90.16 | 150 |
| NFLE | 91.14 | 100 | 92.38 | 140 |
| 2D-PCA | 90.57 | 15 × 48 | 93.15 | 10 × 60 |
| 2D-LDA | 92.18 | 12 × 48 | 93.96 | 13 × 60 |
| BDFLA | 93.48 | 14 × 8 | 95.62 | 15 × 10 |

*Conclusion:* In this Letter, a novel algorithm called BDFLA is proposed for image feature extraction. It is a NFL-based feature extraction algorithm, which aims to minimise the within-class scatter and maximise the between-class scatter based on the 2D-FL metric. Different from the classical NFL-based approaches, the proposed BDFLA is a matrix-based algorithm. The experimental results on the COIL20 database and the FKP database show the effectiveness of the proposed algorithm.

*Acknowledgment:* This work was supported in part by the National Natural Science Foundation of China (nos. 61371178, 61202456 and 70903016).

© The Institution of Engineering and Technology 2015
*10 November 2014*
doi: 10.1049/el.2014.3834

Lijun Yan, Xiaorui Zhu, Jeng-Shyang Pan and Linlin Tang (*Harbin Institute of Technology, Shenzhen Graduate School, Shenzhen, People's Republic of China*)

✉ E-mail: hittang@126.com

Jun-Bao Li (*Harbin Institute of Technology, Harbin, People's Republic of China*)